\documentclass{ws-ijsc}
\usepackage[sort,compress]{cite}
\usepackage{url}
\usepackage{amsmath,amssymb,amsfonts}
\usepackage{algorithmic}
\usepackage{subfig}
\usepackage{graphicx}
\usepackage{booktabs}
\usepackage{textcomp}
\usepackage[table,xcdraw]{xcolor}
\usepackage{listings}
\usepackage{microtype}
\usepackage{comment}
\usepackage{siunitx}

\newcommand{\asFrame}[1]{\mathcal{#1}}
\newcommand{\R}{\mathbb{R}}

\DeclareMathOperator*{\minimize}{min.}

\usepackage[ruled, vlined, linesnumbered, algo2e]{algorithm2e}
\RestyleAlgo{ruled}
\SetKwProg{Init}{\texttt{setupImpl}}{}{}
\SetKwProg{Step}{\texttt{stepImpl}}{}{}
\SetKwProg{getFeetH}{\texttt{Robot.get\_feet\_H}}{}{}
\SetKwProg{getWorldTransform}{\texttt{wrappers.getWorldTransform}}{}{}
\SetKwProg{getWorldTransformNew}{\texttt{SFwrappers.getWorldTransform}}{}{}

\begin{document}

\markboth{Nuno Guedelha, Venus Pasandi, Giuseppe L'Erario, Silvio Traversaro, Daniele Pucci}
{Simulator of Floating-base Robots in Contact with the Ground}

\catchline{0: Ahead of Print}{2024}{1}{17}{10.1142/S1793351X24300036}{}{}
%
%%%%%%%%%%%%%%%%%%%%%%%%%%%%%%%%%%%%%%%%%%%%%%%%%%%%%%%%%%%%%%%%%%%%

\title{A Flexible MATLAB/Simulink Simulator for Robotic Floating-base Systems in Contact with the Ground: \\
Theoretical background and Implementation Details}

\newcounter{affiliation}
\setcounter{affiliation}{1}
\newcounter{coAuthors}
\setcounter{coAuthors}{2}

\author{Nuno Guedelha\footnotemark[\value{affiliation}] { }\footnotemark[\value{coAuthors}]}
\address{
\email{nunoguedelha@gmail.com}
\http{www.researchgate.net/profile/Nuno-Guedelha}
}

\author{Venus Pasandi\footnotemark[\value{affiliation}] { }\footnotemark[\value{coAuthors}]}
\address{
\http{www.researchgate.net/profile/Venus-Pasandi}
}

\author{Giuseppe L'Erario\footnotemark[\value{affiliation}]}
\address{}

\author{Silvio Traversaro\footnotemark[\value{affiliation}]}
\address{}

\author{Daniele Pucci\footnotemark[\value{affiliation}]}
\address{}

\address{\footnotemark[\value{affiliation}]
Artificial and Mechanical Intelligence, \\
Istituto Italiano di Tecnologia, \\
Genova, Italy \\
E-mail: name.surname@iit.it \\
}
\renewcommand{\thefootnote}{\fnsymbol{footnote}}
\footnotetext[\value{coAuthors}]{These two authors contributed equally to this work.}
\renewcommand{\thefootnote}{\alph{footnote}}

\maketitle

\begin{history}
\end{history}

\begin{abstract}
This paper presents an open-source MATLAB/Simulink physics simulator for rigid-body articulated systems, including manipulators and floating-base robots.
Thanks to MATLAB/Simulink features like MATLAB system classes and Simulink function blocks, the presented simulator combines a programmatic and block-based approach, resulting in a flexible design in the sense that different parts, including its physics engine, robot-ground interaction model, and state evolution algorithm are simply accessible and editable.
Moreover, through the use of Simulink dynamic mask blocks, the proposed simulator supports robot models integrating open-chain and closed-chain kinematics with any desired number of links interacting with the ground.
This simulator can also integrate second-order actuator dynamics.
Furthermore, the simulator benefits from a one-line installation and an easy-to-use Simulink interface.
\end{abstract}

\keywords{robotic simulator, floating-base robots, contact model, impact model, MATLAB/Simulink.}

\section{Introduction}

Physics simulators provide a rapid, inexpensive and safe test platform for validating theoretical investigations and improving robots prototype (e.g., robot mechanical and control design).
In this paper, we present an open-source MATLAB/Simulink library for the simulation of robots with rigid bodies, including manipulators and floating-base robots.

Robotic simulators could be defined as a framework with 
(i) physics engine for dynamic modelling, and
(ii) contact detection and model.
Simulators include some functionality features like motion visualization, importing scenes and meshes, low reality gap, low complexity, high reproducibility, simple scenario and environment construction, low resource cost, automation features (i.e., allows automated testing and headless, scripted, or parallel execution), high reliability (i.e., simulator stability, timing, and synchronization), and high interface stability~\cite{afzal2020study}.

Various simulators have been proposed for robotic systems including commercial and open-source simulators \cite{collins2021review}.
Both commercial and open-source simulators mostly support multiple physics engines in the sense that they provide the possibility to switch between multiple specific physics engines.
However, simulators rarely support the possibility of editing their physics engine.
From the contact model perspective, simulators use a variety of models.
In fact, the contact model is especially important in simulating human-robot interactions or the dynamics of free-floating robots, typically legged robots.
Different contact models have various parameters affecting their reality gap and stability.
Usually, in commercial simulators, the contact model cannot be modified by the user, while some of its parameters could be adjusted by the user considering the desired robot model, environment, and motion scenario.
Instead, in open-source simulators, the contact model is mostly available in detail, and also the user has the possibility to modify this model by considering the simulation test features~\cite{acosta2022validating}.
Considering the above explanation and the fact that commercial simulators have high resource costs, open-source simulators are an interesting choice for robotic developers \cite{afzal2020study}.

Robotics researchers and engineers widely use MATLAB/Simulink for the design, simulation, and verification of robotic systems.
In this regard, some open-source robotic simulators based on MATLAB have been presented in the literature, including OpenMAS and FROST.
However, OpenMAS is an agent-oriented, and not a physical, simulator with support for human-environment contacts.
FROST is based on both MATLAB and Mathematica.
To the author's knowledge, an open-source MATLAB/Simulink simulator for floating base robotic systems in contact with the environment is still missing.

In this paper, we present an open-source flexible robotic simulator with an easy-to-use interface for rigid body robots, including manipulators and free-floating robots.
The proposed simulator is a Simulink library that relies on iDynTree which is an open-source multibody dynamics library developed in C++ for free-floating robots \cite{nori2015icub}.
For the simulation of robot-environment interaction, we developed and implement a rigid contact dynamics based on \textit{Gauss principle of least constraint}~\cite{udwadia1992new, azad2019effects}.
The developed contact dynamics benefits two features; stability and simple extension for different environments like inclined ground, moving surface, etc.
We used iDynTree functionalities wrapped by Whole-Body toolbox \cite{FerigoControllers2020} through Simulink Function blocks which improve the whole model execution speed and provide the implementation flexibility to replace iDynTree with the desired physics engine.
The proposed simulator is designed and implemented by the use of MATLAB System classes and the programmatic call to dynamics computation functions.
As a result, details of the proposed simulator logic including the contact detection and model, and the state evolution algorithm is simply accessible and modifiable.
Moreover, thanks to the Dynamic Mask Subsystem feature in Simulink, the proposed simulator can be used for various robot models including robots with open-chain or closed-chain kinematics with any desired number of links interacting with the ground.
Finally, a visualizer that relies on iDynTree is also provided for the visualization of robot motion.
Compared to the existing open-source robotic simulators based on MATLAB, our proposed simulator has the following advantages:
(i) Simulink easy-to-use interface,
(ii) implementation flexibility allowing editing the physics engine and contact dynamics of the proposed simulator,
(iii) capability to simply be integrated into the sensor and motor dynamics.

This paper presents the theoretical background and implementation details of the proposed simulator.
Instead, the validation of various features of proposed simulator as well as its performance is discussed in our previous related work \cite{guedelha2022irc}.
All results shall be made available on GitHub\footnote{\url{https://github.com/ami-iit/paper_guedelha_2022_irc_flexible-matlab-simulink-robots-simulator}}.
For reproducing these results, the simulator packages, as well as their dependencies, can be installed through Conda package manager, as described in the project GitHub repository respective section\footnote{\url{https://github.com/ami-iit/matlab-whole-body-simulator#readme}}.

The rest of the paper is organized as follows.
Section~\ref{sec:background} recalls the rigid body floating-base robotic systems dynamics and the \textit{Gauss principle of least constraint} for the contact dynamics.
Section~\ref{sec:mathematical-methods} introduces the mathematical formulation for dynamics of rigid body floating-base robotic systems with closed-chain kinematics and the contact dynamics developed for modeling the interaction between such robotic system and the environment.
Section~\ref{sec:methods} describes the architecture of the proposed Simulink library and its key elements, as well as used features like Simulink Functions, Dynamic Mask Subsystems and subsystem output buses.
Finally, Section~\ref{sec:conclusion} draws the conclusions.
\section{Background}
\label{sec:background}

\subsection{Dynamics formulation}\label{sec:background:dynamics-model}

We define the equation of motion of a floating base system with $n+1$ bodies connected in an open-chain kinematics by $n$ joints and exchanging pure forces with the environment in $n_c$ points as~\cite[Chap. 3]{traversaro_silvio_2017_3564797}
\begin{equation}
    M(q) \dot{\nu} + C(q, \nu) + G(q) = S \tau  + \sum_{i=1}^{n_c} {J_c}_i^{\top} f_i,
    \label{eq:system_dyn}
\end{equation}
where, representing the inertial frame by $\asFrame{I}$:
\begin{itemize}%[leftmargin=*]
    \item $q \in \R^{n+1}$ is the vector of generalized coordinates, defined as $q=({}^\asFrame{I}p_\asFrame{B}, {}^\asFrame{I}R_\asFrame{B}, s) $, where $s$ are the joint positions, ${}^\asFrame{I}p_\asFrame{B}, {}^\asFrame{I}R_\asFrame{B}$ are the position and the orientation of base frame $\asFrame{B}$ w.r.t $\asFrame{I}$;
    \item ${}^\asFrame{I}H_\asFrame{C} \in \R^{4 \times 4}$ is the homogeneous transformation from frame $\asFrame{C}$ to the inertial frame $\asFrame{I}$;
    \item $\nu \in \R^{n+1}$ is the system velocity, defined as $\nu~=~({}^\asFrame{I}\dot{p}_\asFrame{B}, {}^\asFrame{I}\omega_\asFrame{B}, \dot{s})$, where $\dot{s}$ are the joint velocities and ${}^\asFrame{I}\dot{p}_\asFrame{B}$, $ {}^\asFrame{I}\omega_\asFrame{B}$ are the linear and angular velocity of the base frame $\asFrame{B}$ w.r.t. $\asFrame{I}$, satisfying the relationship $\dot{{}^\asFrame{I}R_\asFrame{B}} = {}^\asFrame{I}\omega_\asFrame{B} \times {}^\asFrame{I}R_\asFrame{B} $; 
    \item $ M \in \mathbb{R}^{(n+1) \times (n+1)}$, $C \in \mathbb{R}^{n+1}$ and $G \in \mathbb{R}^{n+1}$ are the inertia matrix, the vector of Centrifugal and Coriolis effects, and the vector of Gravitational effects, respectively.
    \item $h \in \R^{n+1}$ is the vector of generalised bias forces defined as $h:=C + G$.
    \item $ \tau \in \mathbb{R}^{n}$ is the vector of the actuator forces/torques, and $S$ a selector matrix such that $S \tau = [ 0_{1 \times 6} \ \tau^\top ]^\top$
    \item $J_i(q) \in \R^{3 \times (n+6)}$ is the linear part of the Jacobian mapping the system velocity $\nu$ to the velocity ${}^\asFrame{I}\dot{p}_{\asFrame{C}_i}$ of a frame $\asFrame{C}_i$.
    \item $f_i \in \R^3$ is the $i$th external force applied at the origin of frame $\asFrame{C}_i$ expressed in $\asFrame{I}$.
\end{itemize}

For the sake of readability, the dependency on $q$ and $\nu$ shall be omitted (e.g. $M(q)$ written as $M$).

\subsubsection{Rigid contact model}
\label{subsec:contact-model}
The \textit{Gauss principle of least constraint} says that the motion of a constrained system is such that its acceleration is as close as possible to the one of the unconstrained system, called \textit{free acceleration}, in the sense of kinetic energy \cite{udwadia1992new, azad2019effects}~\hspace{-0.2em}.
We can apply this principle to the computation of the contact forces.\looseness=-1

Let us assume that the only forces acting on the system are the reaction ones, the \textit{free acceleration} is defined as the acceleration of the system with no constraints
\begin{equation}
\dot{\nu} _f := M^{-1} \ (S \tau - h).
\end{equation}

We want to find the acceleration of the constrained system that is closest to the free acceleration, defined in terms of kinetic energy, subject to the contact constraint
\begin{subequations}
\begin{align}
\min _{\dot{\nu}} &\quad  \frac{1}{2} ( \dot{\nu} - \dot{\nu} _f )^\top M \ ( \dot{\nu} - \dot{\nu} _f )  \\
s.t. &\quad J_i \dot{\nu}+\dot{J}_i \nu=0, \quad i \in [1, \dots, n_c],
\end{align}
\label{eq:opti_problem}
\end{subequations}

\noindent{where $J_i \dot{\nu}+\dot{J}_i \nu=0$ express the fact that the acceleration of the contact point $i$ is null.}

The problem \eqref{eq:opti_problem} can be solved using the method of the Lagrange multipliers~\cite{nocedal2006numerical}~\hspace{-0.5em}, where the Lagrangian function is 
\begin{equation}
L(\dot{\nu}, \lambda) =  \frac{1}{2} ( \dot{\nu} - \dot{\nu} _f )^\top M \ ( \dot{\nu} - \dot{\nu} _f )  - \sum_{i=1}^{n_c} \lambda_i^\top (J_i \dot{\nu}+\dot{J}_i \nu).   
\label{eq:lagrangian}
\end{equation}

The Lagrangian multiplier $\lambda_i$ represents the contact forces $f_i$ acting on the point $i$. 
The solution of~\eqref{eq:lagrangian} maximizes the dissipation of the kinetic energy of the system
\begin{equation}
    \minimize_{\lambda} \frac{1}{2} \sum_{i=1}^{n_c} \lambda_i^\top J_i M^{-1} J_i^\top \lambda_i + \lambda_i^\top (J_i \dot{\nu}_f + \dot{J}_i \nu).
    \label{eq:max-kin-energy}
\end{equation}
This formulation allows us to relax the rigid contact assumption and opens up the possibility of simulating contacts with slipping. We can complement the optimization problem~\eqref{eq:max-kin-energy} with three additional constraints, namely:
\begin{itemize}%[leftmargin=*]
\item the forces should lie in a \textit{linearized friction cone}~\cite{stewart2000implicit} ${|\lambda_i^t | \le \mu (\lambda_i^n)}$, where $\lambda_i^t$ and $\lambda_i^n$ represent the components that are tangential and normal to the contact surface and $\mu$ is the friction coefficient;
\item the contact forces should fulfil the complementarity condition $\lambda_i^n \cdot c = 0$, i.e., the normal component of the force should be non-null when the point is in contact~\cite{stewart2000implicit}, i.e., $c=0$, and null when the point is not in contact, i.e., $c=1$;\looseness=-1
\item $\lambda_i^n \ge 0$, i.e., the normal component of the force should always be non-negative.
\end{itemize}

The solution $\lambda$ to the optimization problem represents contact forces $f$ that lie in the friction cone and maximize the dissipation of the kinetic energy. 

\subsubsection{Impact model}
\label{sec:background:impact-model}

The formulation from Section~\ref{subsec:contact-model} allows us to find the contact force so that the point in contact does not accelerate but does not consider the system's discontinuous nature when a collision with the ground occurs.
Hence we need to reason about the impact dynamics. We assume that the collision is inelastic, we use the conservation of the momentum for modelling the impact~\cite{hurmuzlu1994rigid} and derive the formulation that follows.
The time integration of the~\eqref{eq:system_dyn} over the acting time $[t^-, t^+]$, around an instant $t$, of a single impulsive impact force from any given point $i$ making contact with the surface, gives~\cite{kraus1997compliant}
\begin{equation}
\int_{t^{-}}^{t^{+}} (M \dot{\nu} {+} h {-} S \tau {-} J_i ^\top f_i) \ \mathrm{d} t  =  M(\nu^{+} {-} \nu^{-}) {-} J_i^\top \mathbf{F}_i = 0,
\label{eq:impact-eq}
\end{equation}
where $\mathbf{F}_i$ is the $i$th impulse\footnote{The impulse, written in bold, is the product of the impulsive force by the acting time.}, and $\nu^{-}$ and $\nu^{+}$ are the generalized velocity before and after the impact.

We assume that the collision is inelastic hence the point after the collision is at rest, i.e., $\dot{x}_i = J_i\nu^{+} = 0$. This constraint, combined with~\eqref{eq:impact-eq}, acts only at the impact and writes as
\begin{equation}
    \begin{bmatrix}
    M & - J_i^\top \\
    J_i & 0
    \end{bmatrix}
    \begin{bmatrix}
    \nu^{+} \\ \mathbf{F}_i
    \end{bmatrix} =  
    \begin{bmatrix}
    -M \nu ^- \\ 0
    \end{bmatrix},
\end{equation}
whose solution is
\begin{equation}
    \nu ^+ = (I - M^{-1} J_i^\top (J_i M^{-1} J_i^\top)^{-1} J_i) \nu ^{-} = N \nu ^{-},
\end{equation}
where $N$ is the matrix that projects the velocity $\nu^{-}$ onto the dynamically consistent manifold. 

\subsubsection{Integration}

The contact forces $f$ and, in case of an impact, the updated generalized velocity $\nu$, are used in~\eqref{eq:system_dyn} to compute the system acceleration
\begin{equation}
    \dot{\nu} = M^{-1} (S \tau +  \sum _{i=1} ^{n_c} J^\top _i f_i - h)
\end{equation}

This acceleration is integrated to obtain the system's generalized coordinates.
\subsection{Software libraries}
\subsubsection{iDynTree}\label{sec:idyntree}
\textit{iDynTree} is an open-source library for efficient computations of robotic dynamics algorithms for control, estimation, and simulation~\cite{nori2015icub}. It is written in C++ and provides Python and MATLAB bindings. Typically, MATLAB bindings for C/C++ are based on MEX\footnotemark{} (Matlab EXecutable) functions which behave just like any MATLAB script or function and call a subroutine in a C/C++ compiled library.
\textit{iDynTree} can read a \textit{Unified Robot Description Format} (URDF) file that represents the robot model.  
As we explain in the following sections, \textit{iDynTree} provides -- through the intermediate layer constituted by \textit{Whole-Body toolbox} -- the building blocks used to build the proposed Simulink-based multi-body simulator.

iDynTree includes visualization capabilities. We use them, through MATLAB binding, to visualize the robot's behavior during the simulation. 

\newcounter{mexFuncFootnote}
\setcounter{mexFuncFootnote}{\value{footnote}} 
\footnotetext[\value{mexFuncFootnote}]{\url{https://www.mathworks.com/help/matlab/call-mex-functions.html}}

\subsubsection{Whole-Body Toolbox}\label{sec:whole-body-toolbox}
\textit{Whole-Body Toolbox} (WBT) is a software layer that simplifies the design of whole-body robotics algorithms. WBT envelops the capabilities of iDynTree through the \textit{BlockFactory} framework~\cite{FerigoControllers2020}. \textit{BlockFactory} generally allows wrapping of C and C++ algorithms and provides an abstraction layer in which every algorithm constitutes a block embedded in \textit{Simulink} and exchanges data. In such a way, the user can also utilize Simulink's rapid prototyping and visual data analysis abilities.

WBT is mainly used to build Simulink-based whole body controllers conceived to control generic robotics systems~\cite{Nava_etal2016}. We employ WBT to compute the dynamics and kinematics quantities that describe the robot's characteristics and behavior in this work. Further details are in Section~\ref{sec:methods:simulink-function-block}.
\graphicspath{{./figs/MathematicalMethods/}}

\section{Mathematical Methods}\label{sec:mathematical-methods}

In this section, we extend the mathematical formulation of the dynamics and contact models explained in Section \ref{sec:background:dynamics-model} for:
(1) floating-base systems with closed-chain kinematics---in the recent years, various parallel actuation mechanisms have been proposed for robotic systems, which require, for their thorough simulation, to consider the motion and dynamic effects of the parallel mechanism that introduces a closed-chain kinematic system;
(2) sliding behaviour during the impact with the ground---in the impact model explained in Section \ref{sec:background:impact-model}, the contact points are assumed to be at rest after an impact (i.e. when at least one contact point collides with the ground) regardless of the friction effects.
This assumption is conservative for multi contact scenarios, typically a legged robot with multiple feet.

\subsection{Dynamics formulation}

A system with closed-chain kinematics can be considered as a system with open-chain kinematics with some topological constraints.
For this purpose, the closed-chain can be split into two open chains by imaginary breaking the closed chain from a link.
In this way, for a floating-base system with $n_p$ closed chains, \eqref{eq:system_dyn} can be modified as
\begin{equation}\label{eq:theory:EOM-closed-chain}
M \dot{\nu} +
h = S \tau  + J_c^\top f_c + \tilde{J}_p^\top \hat{f}_p,
\end{equation}
where:
\begin{itemize}%[leftmargin=*]
    \item $f_c \in R^{3n_c}$ is the grouped contact forces vector, defined as $f_c = \begin{bmatrix} f_1^\top , & f_2^\top & \dots & f_{n_c}^\top \end{bmatrix}^\top$.
    \item $\hat{f}_p \in R^{3n_p}$ is the grouped imaginary external forces vector, defined as $\hat{f}_p = \begin{bmatrix} \hat{f}_1^\top , & \hat{f}_2^\top & \dots & \hat{f}_{n_p}^\top \end{bmatrix}^\top$ where $\hat{f}_i \in \R^3$ is the imaginary external force applied at the origin of the frame $\asFrame{C}_{i1}$ expressed in $\asFrame{I}$.
    \item $\tilde{J}_c \in R^{(3n_c)\times (n+6)}$ is the grouped Jacobian matrix for contact points, defined as $\tilde{J}_c = \begin{bmatrix} J_1^\top, J_2^\top, \dots, J_{n_c}^\top \end{bmatrix}^\top$.
    \item $\tilde{J}_p \in R^{(3n_p)\times (n+6)}$ is the grouped Jacobian matrix for imaginary breaking points, defined as $\tilde{J}_p = \begin{bmatrix} \tilde{J}_1^\top, \tilde{J}_2^\top, \dots, \tilde{J}_{n_p}^\top \end{bmatrix}^\top$ where $\tilde{J}_j(q) \in \R^{3\times (n+6)}$ is the Jacobian difference in the imaginary breaking point $j$, defined as $\tilde{J}_j = J_{j1} - J_{j2}$, where $J_{j1}$ and $J_{j2}$ are the linear part of the Jacobian for the two imprints of the imaginary breaking point.
\end{itemize}
The topological constraint for the closed chain $j$ can be written as
\begin{equation}\label{eq:theory:topo-cnstr:pos}
    p_{j1} - p_{j2} = 0,
\end{equation}
where $p_{k} \in R^{3}$ is the position of origin of frame $\asFrame{C}_{k}$.
The frames $\asFrame{C}_{j1}, \asFrame{C}_{j2}$ are the frames attached to the two imprints of imaginary breaking point $j$.
Using the Jacobian concept and considering the fact that the topological constraint is initially satisfied at $t=0$, the above constraint can be expressed in the acceleration level as
\begin{equation}
    \tilde{J}_j \dot{\nu} + \dot{\tilde{J}}_j \nu = 0.
\end{equation}
In the contact and impact models, we discuss how we compute $\hat{f}$ for ensuring the above constraint.

\subsection{Rigid contact model}

Following the Gauss principle of least constraint, described in Section \ref{subsec:contact-model}, the acceleration of the constrained system with closed-chain kinematics is the closest value to the system free acceleration subject to the topological and contact constraints as
\begin{subequations}
\begin{align}
\min _{\dot{\nu}} &\quad  \frac{1}{2} ( \dot{\nu} - \dot{\nu} _f )^\top M \ ( \dot{\nu} - \dot{\nu} _f )  \\
s.t. &\quad \tilde{J}_p \dot{\nu} + \dot{\tilde{J}}_p \nu = 0, \\
&\quad J_c \dot{\nu} + \frac{J_c}{\Delta t} \nu + \dot{J}_c \nu=0 \label{eq:theory:contact_cnstr},
\end{align}
\label{eq:opti_problem_2}
\end{subequations}
where $\Delta t \in R$ is the Euler integration time step. \eqref{eq:theory:contact_cnstr} expresses the fact that the velocity of contact points is null.
Using KKT conditions, we obtain
\begin{equation}\label{eq:theory:KKT-conditions}
\begin{aligned}
    &\tilde{J}_p M^{-1} \left( \tilde{J}_p^\top \hat{f}_p + J_c^\top f_c \right) + \tilde{J}_p \dot{\nu}_f + \dot{\tilde{J}}_p \nu = 0, \\
    &J_c M^{-1} \left( \tilde{J}_p^\top \hat{f}_p + J_c^\top f_c \right)  + J_c \dot{\nu}_f + \frac{J_c}{\Delta t} \nu + \dot{J}_c \nu = 0.
\end{aligned}
\end{equation}
Considering the first equation, we have
\begin{equation}
    f_p = - \left( \tilde{J}_p M^{-1} \tilde{J}_p^\top \right)^{-1} \left( \dot{\tilde{J}}_p \nu + \tilde{J}_p \dot{\nu}_f + \tilde{J}_p M^{-1} J_c^\top f_c \right),
\end{equation}
and substituting $f_p$ into the second one, we obtain
\begin{equation}
    H f_c + g = 0,
\end{equation}
where
\begin{equation}
    \begin{aligned}
        H = &J_c M^{-1} J_c^\top - J_c M^{-1}\tilde{J}_p^\top\\
        & \left( \tilde{J}_p M^{-1} \tilde{J}_p^\top \right)^{-1} \tilde{J}_p M^{-1} J_c^\top, \\
        g =& \frac{J_c}{\Delta t} \nu + \dot{J}_c \nu + J_c \dot{\nu}_f - J_c M^{-1} \tilde{J}_p^\top \\
        &\left( \tilde{J}_p M^{-1} \tilde{J}_p^\top \right)^{-1} \left( \dot{\tilde{J}}_p \nu + \tilde{J}_p \dot{\nu}_f\right).
    \end{aligned}
\end{equation}
The solution of the above equation maximizes the dissipation of the kinetic energy of the system
\begin{equation}
    \minimize_{f_c} \frac{1}{2} f_c^\top H f_c + f_c^\top g.
    \label{eq:max-kin-energy-2}
\end{equation}
Thanks to the above formulation, we are able to relax rigid contact assumption by complementing the above optimization problem with the 3 friction constraints stated in Section \ref{subsec:contact-model}.

\subsection{Impact model}

The time integration of \eqref{eq:theory:EOM-closed-chain} over the acting time of the impulsive impact gives
\begin{equation}\label{eq:theory:integrated-EOM}
    M \left( \nu^+ - \nu^- \right) = J_c^\top \mathbf{F}_c + \tilde{J}_p^\top \mathbf{F}_p, 
\end{equation}
where $\mathbf{F}_c, \mathbf{F}_p$ are the grouped vector of the impulsive contact and imaginary external forces.
The above equation is subjected to the topological and contact constraints as
\begin{equation}
    \begin{aligned}
        \tilde{J}_p \nu^+ = 0, \\
        J_c \nu^+ = 0.
    \end{aligned}
\end{equation}
Note that there is also two constraints on $\nu^-$, but since $\nu^-$ is a known input we don't need to include it in the impact model for computing $\nu^+$.
Substituting $\nu^+$ from \eqref{eq:theory:integrated-EOM} into the topological constraint, we have
\begin{equation}
    \mathbf{F}_p = -\left( \tilde{J}_p M^{-1} \tilde{J}_p^\top \right)^{-1} \left( \tilde{J}_p M^{-1} J_c^\top \mathbf{F}_c +\tilde{J}_p \nu^- \right).
\end{equation}
Substituting $\nu^+$ and $\mathbf{F}_p$ into the contact constraint, we have
\begin{equation}
    H \mathbf{F}_c + g = 0,
\end{equation}
where
\begin{equation}
    \begin{aligned}
        &H = J_c M^{-1} \left(I - \tilde{J}_p^\top \left( \tilde{J}_p M^{-1} \tilde{J}_p^\top \right)^{-1} \tilde{J}_p M^{-1} \right) J_c^\top, \\
        &g = J_c \left(I - M^{-1} \tilde{J}_p^\top \left( \tilde{J}_p M^{-1} \tilde{J}_p^\top \right)^{-1} \tilde{J}_p \right).
    \end{aligned}
\end{equation}
The solution of above equation can be expressed as
\begin{equation}
    \mathbf{F}_c = \arg\minimize_{\mathbf{F}_c} \frac{1}{2} \mathbf{F}_c^\top H \mathbf{F}_c + \mathbf{F}_c^\top g.
\end{equation}
This formulation allows us to relax the rigid contact assumption and considering the possibility of the motion of the contact points.
For this purpose, we complement the above optimization problem with the three friction constraints explained in Section \ref{subsec:contact-model}.
\graphicspath{{./figs/Methods/}}

\section{Methods}\label{sec:methods}

\subsection{Overview of the simulator}\label{sec:methods:overview-of-the-simulator}

\begin{figure}[t]
    \centering
        \includegraphics[width=0.45\textwidth]{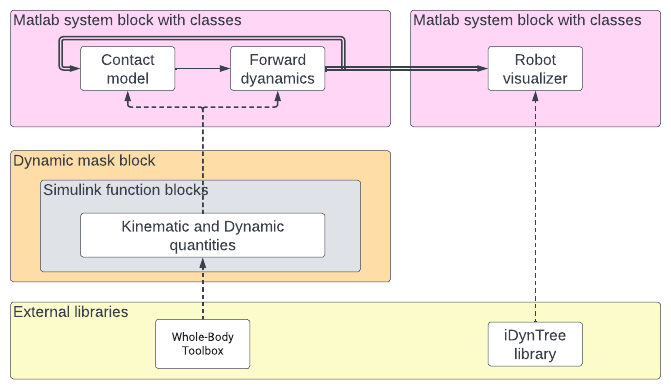}
    \caption{Overview of the Simulator. Solid, single and double arrows represent the data flow between the different blocks, double arrows being used for buses. Dashed arrows denote the programmatic call of its source block.}
    \label{fig:simulator-overview}
\end{figure}

As shown in Figure \ref{fig:simulator-overview}, the Simulator core consists of three main blocks, namely two MATLAB System blocks and a set of Simulink Function blocks redefined as Dynamic Mask Subsystems.
The first MATLAB System Block contains a MATLAB class for computing the time evolution of the robot states.
For this purpose, the contact model computes the reaction forces applied to the robot by the ground, and a forward dynamic model computes the robot acceleration considering the applied forces.
For both contact and forward dynamic models, we need some kinematic and dynamic quantities of the robot, like link Jacobian matrices, inertia matrix, Coriolis and Centrifugal effects vector, etc.
For computing the kinematic and dynamic quantities, we use Simulink Functions which call different functionalities of the Whole-Body Toolbox (see Section~\ref{sec:whole-body-toolbox}).
The second MATLAB System block contains a MATLAB class for visualizing the robot's motion.
For the visualization, we use the iDynTree library (see Section~\ref{sec:idyntree}).
In the following sections, we explain in detail the simulator elements described above, focusing on the implementation of the first MATLAB System and the Dynamic Mask Subsystem blocks.

\subsection{MATLAB System Block with classes}\label{sec:methods:matlab-system-block-with-classes}

MATLAB System blocks allow the object-oriented implementation of complex specific algorithms, which typically handle rigid unilateral contacts, second-order dynamics or joint limits, and their use in Simulink~\cite{MatlabSystemBlockWebPage}.

They can be used to simulate a dynamical system's state evolution with time-varying inputs. Two main methods of this block are the \texttt{setupImpl} method, in which attributes are initialized, and classes are instantiated, and the \texttt{stepImpl}, in which the algorithm is implemented (the same logic applies to the \texttt{Robot visualizer}).

The behavior of the MATLAB System block simulating the robot state evolution (top left MATLAB System block in Figure \ref{fig:simulator-overview}) depends on the current input and its state (the joint torques $\tau$ and the robot's configuration and velocity $q=({}^\asFrame{I}p_\asFrame{B}, {}^\asFrame{I}R_\asFrame{B}, s) $, $\nu~=~({}^\asFrame{I}\dot{p}_\asFrame{B}, {}^\asFrame{I}\omega_\asFrame{B}, \dot{s})$, respectively) stored in the internal attributes of the class.\looseness=-1
We implemented this behavior through three pure MATLAB classes: the class \textbf{Robot} computes the forward dynamics of the robot, namely the configuration acceleration driven by the joint torques; the class \textbf{Contact} computes the reaction forces; the class \textbf{State} computes the system's state evolution.
The robot acceleration is integrated over the integration step.

Instances of these three classes are initialized in the \texttt{setupImpl} method while the \texttt{stepImpl} method implements the algorithm, as shown in Algorithm \ref{algo:robot_dyn_with_contacts}. The robot state ($q$, $\nu$), updated at each step, is exposed on the MATLAB System output ports.

\begin{algorithm2e}[H]
\DontPrintSemicolon
\SetKwInOut{Input}{input}\SetKwInOut{Output}{output}\SetKwInOut{State}{internal state}
\Input{joint torques $\tau$}
\Output{contact wrenches $f$}
\State{robot state $q$, $\nu$}
\Init{}{initialize \texttt{Robot}\;
        initialize \texttt{Contact}\;
        initialize \texttt{State}\;}
\While{simulation is not over}
{\Step{}{
    $\mathbf{f} \leftarrow 0 $ \;
    \uIf{robot in contact}{   
        $f, \nu \leftarrow \text{compute\_contacts}(q, \nu, \tau)$ \;
    }
    $\dot{\nu} \leftarrow \text{forward\_dynamics}(q, \nu, \tau, f)$ \;
    $q, \nu \leftarrow \text{integrate}(\dot{\nu})$\;
    \Return{$f$}
}}
\caption{Robot dynamics with contacts}
\label{algo:robot_dyn_with_contacts}
\end{algorithm2e}

The same logic applies to the \texttt{Robot visualizer} (top right MATLAB System block in Figure \ref{fig:simulator-overview}). In the \texttt{setupImpl} an instance of the iDynTree visualizer is initialized along with the environment, while in the \texttt{stepImpl} the robot configuration $q$ is set to the robot and hence visualized.

\subsection{Simulink Function Block}\label{sec:methods:simulink-function-block}

The motivation for using Simulink Function blocks was to maintain a good execution performance while taking advantage of the implementation flexibility brought by the programmatic call to external dynamics libraries.

The design and implementation of complex algorithms were made easier by the use of MATLAB System classes and the programmatic call, through binding MEX functions, to external dynamics computation libraries like iDynTree.
Such an improvement came with a downside, as it significantly slowed down the simulation, lowering the real-time factor.
This was due to the model being simulated via the MATLAB interpreter engine which is quite slow compared to Simulink's code generation\footnote{https://www.mathworks.com/help/simulink/ug/simulation-modes.html}.
The latter supports only a subset of MATLAB functions and is not compatible with direct calls to MEX functions\footnotemark[\value{mexFuncFootnote}].
Simulink Functions allow replacing the former bindings with programmatic calls to custom, self-contained Simulink blocks, like the WBT (Whole-Body Toolbox) blocks which implement a C++ abstraction layer wrapping the iDynTree library through the \textit{BlockFactory} framework~\cite{FerigoControllers2020}.
The WBT blocks are code generation ready, allowing a significant execution speed improvement~\cite{guedelha2022irc}~\hspace{-0.2em}, while keeping a compact and flexible implementation of complex algorithms, and a flexible analysis capability as it allows users to use Simulink's rapid prototyping and visual data analysis abilities.

\subsubsection*{Integrating a Simulink Function example: \texttt{simFunc\_qpOASES}}

We've created Simulink Functions which wrap WBT blocks implementing dynamics computations methods from the iDynTree library and a multiple QP (Quadratic Programming) solvers.

\begin{figure}[t]
        \hspace*{-0.6cm}\includegraphics[width=\columnwidth]{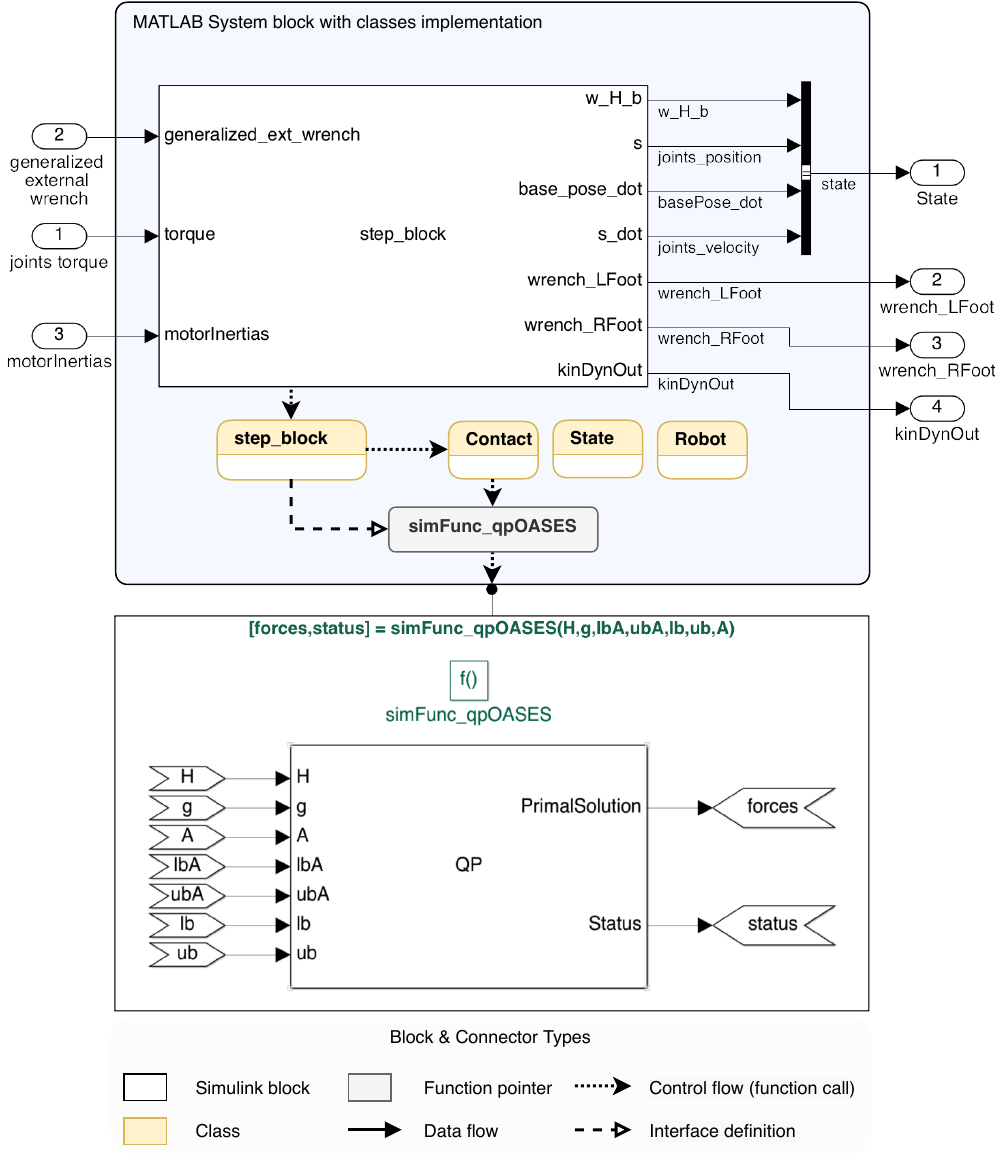}
    \caption{Control flow between the MATLAB System with classes and the Simulink Function wrapping the QP solver.}
    \label{fig:simFunc-qpOASES-control-flow-reduced}
\end{figure}

For understanding how an external library method is integrated in our main model via a Simulink Function, we consider the function \texttt{simFunc\_qpOASES} depicted in Figure \ref{fig:simFunc-qpOASES-control-flow-reduced}.

The SF (Simulink Function) wraps a WBT block "QP" and is triggered by a MATLAB function call in the class \texttt{Contact} implementing the contact model algorithm.

The SF sub-system block mask displays an editable tag defining the function prototype, which matches the calling MATLAB function prototype declared in the MATLAB System interface.
The SF sub-system input and output ports, mapping the SF function respective input and output arguments, are spawned inside the sub-system and connected to the WBT block.
The latter wraps a QP (Quadratic Programming) solver from the qpOASES library\footnote{https://github.com/coin-or/qpOASES}.

We can see in Figure \ref{fig:simFunc-qpOASES-control-flow-reduced} the control flow from the calling MATLAB System to the QP solver.
This model scales to the full set of Simulink Functions, configured as Dynamic Mask Subsystems (refer to Section \ref{sec:methods:dynamic-mask-block}), which implement the kinematics and dynamics computations invoked by the contact model and forward dynamics algorithms.

Another example, depicted in Table \ref{tab:criticalFunctionsForPerfComparison}, illustrating how the execution of multiple intermediate MEX functions is replaced by a single Simulink Function call, reducing significantly the processing time~\cite{guedelha2022irc}.

\subsubsection*{Choosing different function implementations}

It's common to try multiple implementations of a QP problem solver, depending on their performance and stability.
Our simulator integrates three QP solvers: the native MATLAB solver \texttt{quadprog}~\footnote{\url{https://www.mathworks.com/help/optim/ug/quadprog.html}}; OSQP~\footnote{\url{https://osqp.org/docs/index.html}} (Operator Splitting solver for Quadratic Programs)~\cite{osqp}~; and qpOASES~\footnotemark[6], an open-source C++ implementation of an online active set strategy, mentioned in the previous sections. the last two solvers are each wrapped in a WBT block. We can select the solver through a variable in the \texttt{Contact} class.

\begin{table*}[t]
\centering
\caption{Example of a critical $\texttt{Robot}$ class method using Simulink Functions instead of MEX functions.}
\label{tab:criticalFunctionsForPerfComparison}
\resizebox{\textwidth}{!}{%
\begin{tabular}{@{}lll@{}}
\toprule
\rowcolor[HTML]{ECF4FF} 
\multicolumn{1}{c}{\cellcolor[HTML]{ECF4FF}\begin{tabular}[c]{@{}c@{}}Dynamics\\ Functions\end{tabular}} &
  \multicolumn{1}{c}{\cellcolor[HTML]{ECF4FF}Implementation with bindings} &
  Implementation with Simulink Functions \\
  \midrule
  \textbf{ForwardKinematics} &
  \begin{tabular}[c]{@{}l@{}}get\_feet\_H(obj)\\   $\hookrightarrow$ wrapper.getWorldTransform(Lfoot\_frame\_name)\\          $\hookrightarrow$ iD.KinDynComputations.getWorldTransform\\                   $\hspace{5mm} \hookrightarrow$ iDynTreeMEX(\textless{}index\_n\textgreater{}) $\rightarrow$MEX exec\\          $\hookrightarrow$ iD.Transform.asHomogeneousTransform\\                   $\hspace{5mm} \hookrightarrow$ iDynTreeMEX(\textless{}index\_n\textgreater{}) $\rightarrow$MEX exec\\          $\hookrightarrow$ iD.Matrix4x4.toMatlab\\                   $\hspace{5mm} \hookrightarrow$ iDynTreeMEX(\textless{}index\_n\textgreater{}) $\rightarrow$MEX exec\\   $\hookrightarrow$ wrapper.getWorldTransform(Rfoot\_frame\_name)\end{tabular} &
  \begin{tabular}[c]{@{}l@{}}get\_feet\_H(obj)\\   $\hookrightarrow$ SF.KinDynComputations.getWorldTransformLRfoot\\        $\hspace{5mm} \hookrightarrow$ simFunc\_getWorldTransformLFoot $\rightarrow$ Simulink block\\ \\ \\ \\ \\ \\        $\hspace{5mm} \hookrightarrow$ simFunc\_getWorldTransformRFoot $\rightarrow$ Simulink block\end{tabular} \\
  \botrule
\end{tabular}%
}
\end{table*}

\subsection{Output bus with multiple numerical signals}\label{sec:methods:output-bus-with-multiple-numerical-sugnals}

\subsubsection{Motivation} We can wrap the output signals of a MATLAB System block in a bus port~\cite{UsingBusesWithMatlabSystemBlocksWebPage} by setting an interface bus programmatically~\cite{SimulinkBusSpecifyPropertiesOfBusesWebPage}, thus avoiding the usage of additional Simulink ``Signal Routing" elements like a ``Bus Creator". The same can be done on input signals, in order to:
reduce the clutter in a model growing in complexity;
directly interconnect multiple MATLAB System blocks and move them around without caring about re-arranging signal connection lines.

\subsubsection{Implementation} We integrated such a bus output port in the first MATLAB System block (introduced in \ref{sec:methods:overview-of-the-simulator}) implemented by the binded class \texttt{step\_block} which computes the system state evolution algorithm.
The \texttt{kynDynOut} bus wraps all the kinetic and dynamic quantities computed by that algorithm at each simulation time step: the robot state and Jacobian matrices; the inertia matrix; the generalized bias forces and so on.

\subsubsection{Defining the bus structure} Each output variable of a MATLAB System block has to be fully described through a set of properties: name, dimensions, mode, sample time, and so on. In the case of a bus, this description is scaled up to the set of signals contained in the bus, mapped into a \texttt{Simulink.Bus} object which is instantiated in the global workspace or in a dictionary~\cite{SimulinkBusSpecifyPropertiesOfBusesWebPage}~\hspace{-0.2em}~\cite{SpecifyBusPropertiesWithSimulinkBusObjectDataTypesWebPage}~\hspace{-0.2em}.

\subsection{Dynamic Mask Subsystem}\label{sec:methods:dynamic-mask-block}

\begin{figure}[t]
    \centering
    \subfloat[Two feet robot]{
        \includegraphics[width=0.45\textwidth]{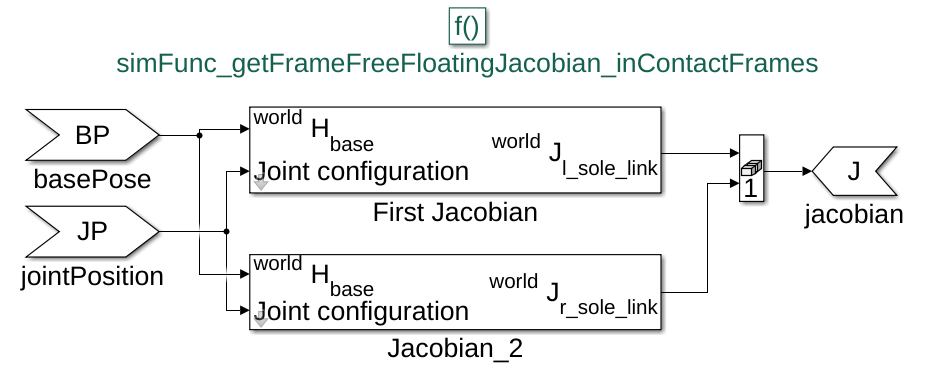}
    }
    \subfloat[Four feet robot]{
        \includegraphics[width=0.45\textwidth]{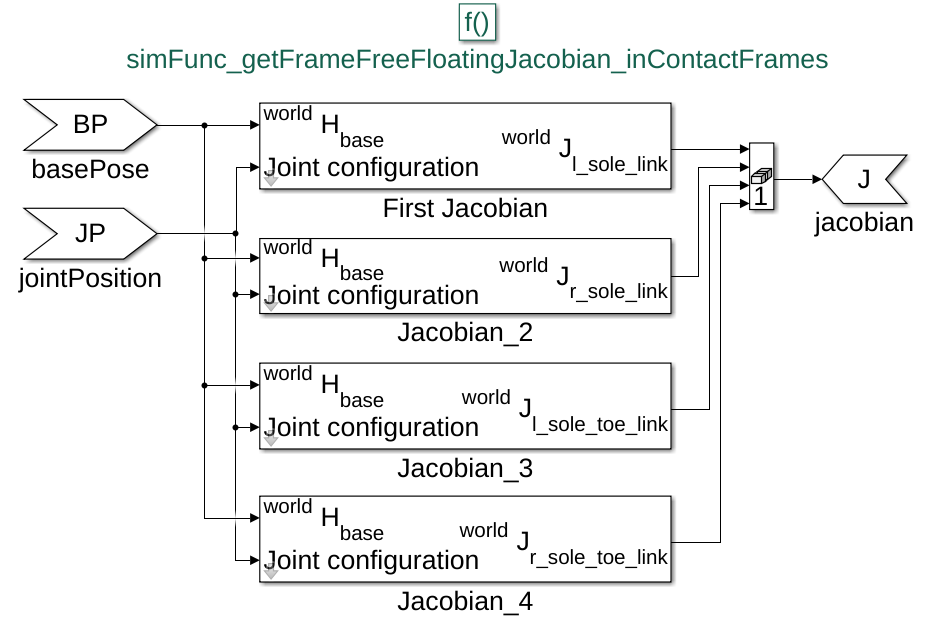}
    }
    \caption{Simulink Function block for computing the Jacobian of the feet for two robots having two and four feet.}
    \label{fig:dynamic-mask-block:feet-jacobian}
\end{figure}

Various robot models have different kinematic and dynamic characteristics, like a different number of feet\footnote{we define feet as the links that interact with the ground}, and open-chain and closed-chain kinematics with a different number of closed-chains.
The simulation of these different features requires different algorithms and information.
To provide those, we sometimes need to modify the content of some subsystem blocks in the Simulink model, typically for adapting that content to the desired robot model.
That can be simply done by redefining the subsystem block as a Dynamic Mask Subsystem~\cite{MatlabDynamicMaskedSubsystemWebPage}.
This step consists in changing its mask configuration by adding a mask-initialization code meant to modify the subsystem content depending on the model configuration parameters.
For example, we need the Jacobian of the feet frame for the forward dynamic computations.
As explained in \ref{sec:methods:simulink-function-block}, the Jacobian matrix is computed by a Simulink Function block using the WBT Jacobian block.
The required number of WBT Jacobian blocks wrapped in the Simulink Function block depends on the number of feet defined for the robot, e.g. two or four as shown in Figure \ref{fig:dynamic-mask-block:feet-jacobian}. We define the function as a Dynamic Mask Subsystem for adding dynamically the desired number of WBT blocks.

\section{Conclusion}\label{sec:conclusion}

In this work, we have presented an open-source MATLAB/Simulink simulator for rigid body systems interacting with the ground. It provides an easy-to-use interface for plugging-in and parameterizing manipulators, free-floating robots, e.g.: humanoid robots; robot models with open-chain or closed-chain kinematics.
The contact model relies on the \textit{Gauss principle of least constraint}, which allows modelling a rigid contact interaction with the environment, properly handling the friction and eventual sliding behaviour during the impact with the ground.

Beyond these features, the initial motivation for developing such a simulator was to improve the flexibility in the implementation of dynamics computations algorithms or physics models while keeping acceptable simulation performances compared to reference available simulators like Gazebo.

We achieved that goal by using object-oriented MATLAB features, like classes, and MEX-based bindings to C++ dynamics algorithms in the iDynTree external library.
This approach turned to be very costly for the execution speed performance, as shown in the MATLAB/Simulink profiling done in our previous related work~\cite{guedelha2022irc}.
We tackled the problem by dropping pure bindings, which allowed code generation and replacing them with programmatic calls to native or custom Block Factory based Simulink blocks. This combined block-based and programmatic approach improved the design flexibility while matching comparable execution speed performance met with a reference simulator like Gazebo.

Thus, the benefit of the proposed library lies in adding to the intrinsic Simulink rapid prototyping, visual data analysis and profiling features, the possibility of: quickly editing and adding sensors and motor dynamics, growing away from simplified to more complex actuator and friction models; changing seamlessly simulation requirements and model parameters through Dynamic Mask Subsystems and bus data exchange.

\section*{ORCID}

\begin{tabular}{l l}
Nuno Guedelha & \url{https://orcid.org/0000-0002-1117-2428} \\
Venus Pasandi & \url{https://orcid.org/0000-0002-1671-0547} \\
Giuseppe L’Erario & \url{https://orcid.org/0000-0001-6042-3222} \\
Silvio Traversaro & \url{https://orcid.org/0000-0002-9283-6133} \\
Daniele Pucci & \url{https://orcid.org/0000-0002-7600-3203}
\end{tabular}

\bibliographystyle{ws-ijsc}
\bibliography{main}

\end{document}